\documentclass{article}

\usepackage{arxiv}

\usepackage[utf8]{inputenc} % allow utf-8 input
\usepackage[T1]{fontenc}    % use 8-bit T1 fonts
\usepackage{hyperref}       % hyperlinks
\usepackage{url}            % simple URL typesetting
\usepackage{booktabs}       % professional-quality tables
\usepackage{amsfonts}       % blackboard math symbols
\usepackage{nicefrac}       % compact symbols for 1/2, etc.
\usepackage{microtype}      % microtypography
\usepackage{lipsum}
\usepackage{graphicx}
\usepackage{amsmath}
\usepackage{multirow}  

\title{Predict future sale}

\title{New complex network building methodology for High Level Classification based on attribute-attribute interaction}

\author{Esteban Wilfredo Vilca Zuñiga\\
Dept. of Computing and Mathematics\\
FFCLRP-USP\\
Ribeirão Preto, Brasil\\
\texttt{evilcazu@usp.br}}

\begin{document}
\maketitle
\begin{abstract}
High-level classification algorithms focus on the interactions between instances. These produce a new form to evaluate and classify data. In this process, the core is the complex network building methodology because it determines the metrics to be used for classification. The current methodologies use variations of kNN to produce these graphs. However, this technique ignores some hidden pattern between attributes and require normalization to be accurate. In this paper, we propose a new methodology for network building based on attribute-attribute interactions that do not require normalization and capture the hidden patterns of the attributes. The current results show us that could be used to improve some current high-level techniques.
\end{abstract}

% keywords can be removed

\section{Introduction}

The machine learning classification algorithms are low level when they use just physical features to classify usually distance measures like euclidean distance. However, high-level classification algorithms focus on the interaction between the data. Using metrics that evaluate the behavior of each node concerning others \cite{zhao2016mlcn}.

These interactions between instances are usually represented as complex networks. There is a variety of techniques to build networks but usually, they use kNN as the core \cite{carneiro2018importanceconcept} \cite{tiago2018hldc} \cite{seyed2019}. These metrics produce a network where each node represents an instance and each edge represents a neighbor in kNN.

A complex network is defined as a non-trivial graph \cite{barabasi2002smcn}. Usually, the quantity of instances on the dataset and the interactions generates a large graph with numerous edges. This large graphs presents special characteristics that are exploited in many techniques to classify data like Betweenness Centrality, Clustering Coefficient, Assortativity, HLNB\_BC and so on.

In order to produce the best classification, we need to present structures that capture all the interactions in the dataset. The current techniques are based on kNN that exploits the relationship between instances. Nevertheless, they just capture instance-instance interactions but there are some hidden patterns on attribute-attribute interaction that are omitted.  

In this paper, we will present a new methodology that captures these attribute-attribute interactions and how the current high-level prediction techniques are affected by this new paradigm.  

\section{Current Methodologies review}

A complex network presents two main parts, the nodes $\mathcal{V}$ and the links $\mathcal{E}$. 
Many methodologies are based on kNN to build the graphs. They use each instance $X_i$ as a node $V_i$ and the links are the connections between this node and its k neighbors. These neighbors in the graph are represented as the neighborhood $\mathcal{N}$ of each node. 
If we are going to use the network for supervised learning, we will need to add the label $y_i$ as a parameter to remove all the links were the neighbors labels are different from the evaluated node \cite{thiago2012hldc}.

\begin{equation} \label{KNN_network_construction_rule}
  \mathcal{N}(X_i)=
    kNN(X_i,y_i)
\end{equation}
Where $X_i$ is the instance and $y_i$ is the label of the instance.
The neighborhood $\mathcal{N}$ will give us the nodes connected to the main node $V_i$ following the next rule $\{ V_j, (V_i,V_j)\in \mathcal{V} : V_j \in \mathcal{N}(X_i) \}$.
When the graph is sparse this methodology captures good relations, but in dense graphs ignore close relationships. There is another variation adding $\epsilon\text{-}radius$ algorithm to capture this dense regions \cite{thiago2012hldc}.

\begin{equation} \label{radius_network_construction_rule}
  \mathcal{N}(X_i)=\begin{cases}
    kNN(X_i,y_i), & \text{otherwise}\\
    \epsilon\text{-}radius(X_i,y_i), & \text{if }|\epsilon\text{-}radius(X_i,y_i)|>k\\
  \end{cases}
\end{equation}

Where $\epsilon\text{-}radius(X_i,y_i)$ returns the set of nodes $\{ V_j, V_j \in \mathcal{V} : distance(X_i,X_j)< \epsilon \land y_i = y_j  \}$. The $distance$ measure could be a similarity function like euclidean distance. The $\epsilon$ value is determinated according to the sparsity of the network. A common value is the median value of the $kNN_{distances}$  \cite{tiago2018hldc}.

This methodology need the normalization of the data to be used by current high-level prediction methodologies and considering a node $V_i$ as an entire instance $X_i$ ignores some hidden-patterns between attributes. 

\subsection{Hidden Patterns}
There are some problems related to the core of these methodologies.

\subsubsection{Ignoring attribute-attribute hidden sub-patterns}

\begin{figure}[h]
\label{fig:twoLinkedNodes}
    \centering
   	\includegraphics[height=2.5cm]{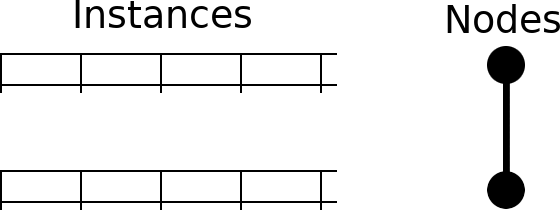}
   	\caption{Image of two related instances transformed in two linked nodes.}
\end{figure}

In figure \ref{fig:twoLinkedNodes}, we can observe how two instance are reduced to just two nodes omitting possible patterns in the same attribute.

\subsubsection{Robustness reduced for normalization}

Due to the reduction of the attributes in one instance, we need normalization because each attribute could present different scale values.

\subsubsection{Non connected graphs}

By cause of the reduction of dimensions, if we do not have a high number of neighbors the networks could be disconnected. Some methodologies must introduce an extra node class to avoid this problem like \textit{HLNB\_BC classification} technique.
\section{Methodology description}
\begin{figure}[h]
    \centering
    \includegraphics[height=4cm]{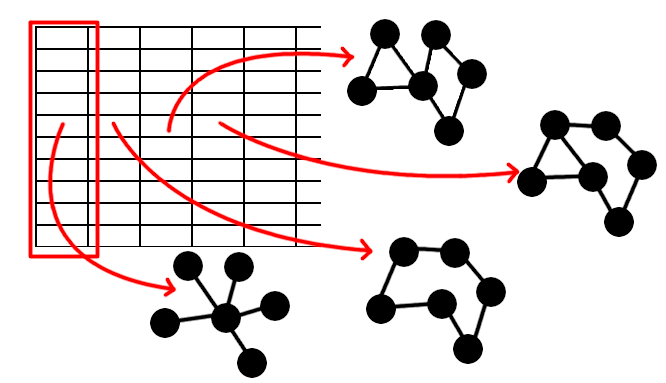}
    \caption{Image with of dataset with 4 graph representation for each attribute.}
    \label{fig:4attributes}
\end{figure}

Our methodology, has three main parts.
First, we will use each attribute-attribute interaction as an independent network like in figure \ref{fig:4attributes}. As a result, we will capture possible hidden patterns for each attribute. Also, this will reduce the normalization dependence because each attribute present the same scale. These networks will follow the same neighbor hood equation (\ref{KNN_network_construction_rule}) in one dimension.

Second, we will combine all the current networks following the same node index in each network. Hence, each node will become in a meta node that absorbs all the links from each attribute. Then, we will use the current techniques like the equation (\ref{radius_network_construction_rule}) for these meta nodes to preserve the instance-instance interaction.

Third, this combined graph could be used for current high-level classification techniques. The new testing instances inserted must follow the same process. It has to be divided for each attribute, combine in a unique node, and use this meta node in the equation  (\ref{radius_network_construction_rule}).

\begin{figure}[h]
        \centering
        \includegraphics[height=4.5cm]{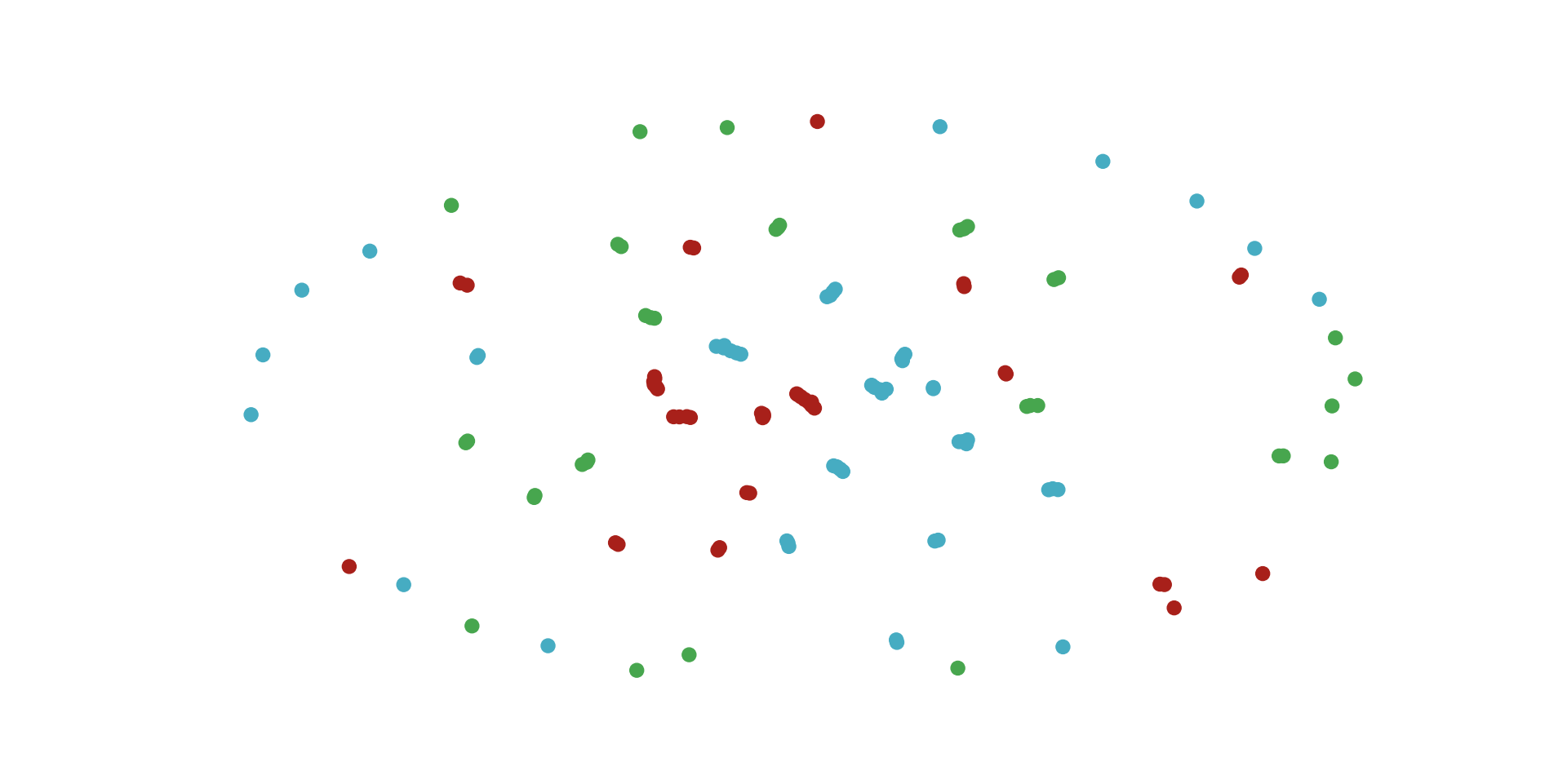}
        \caption{Image of disconnected complex network using the current techniques, kNN (k=1) and $\epsilon$-radius ($\epsilon$ = median of $kNN_{distances}$) on Wine UCI Dataset with three attributes red, blue and green nodes.}
        \label{fig:wineCurrent}
    \end{figure}

\begin{figure}[h]
        \centering
        \includegraphics[height=4.5cm]{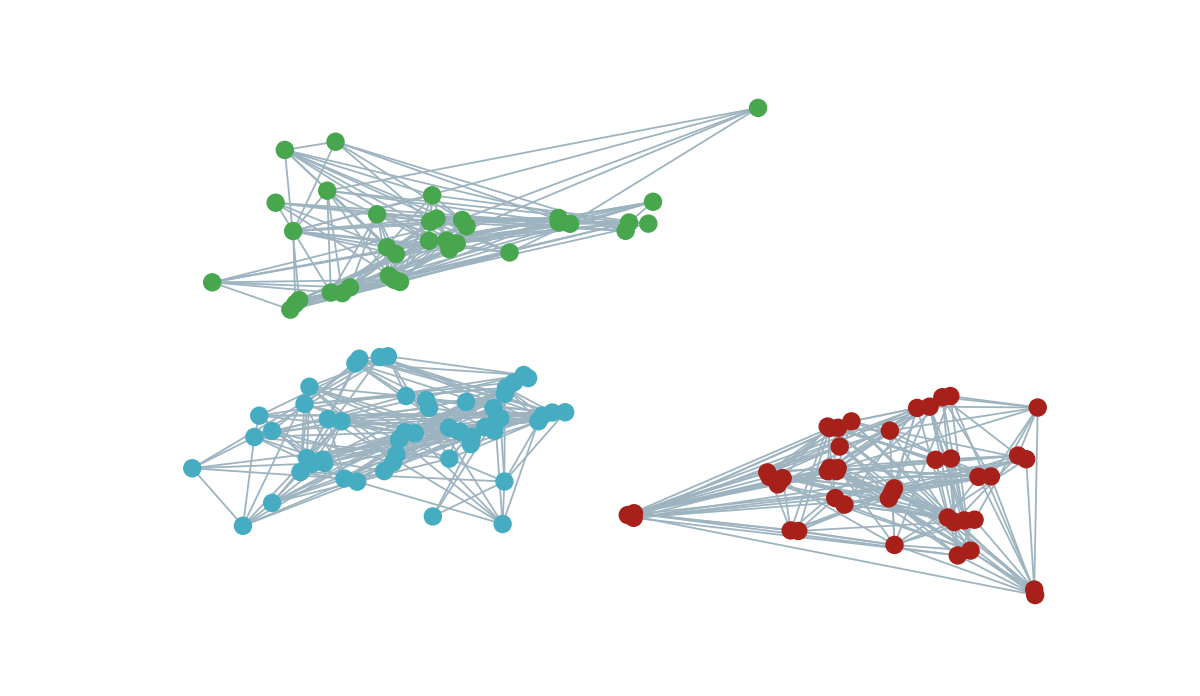}
        \caption{Image of complex network using our methodology, kNN (k=1) and $\epsilon$-radius ($\epsilon$ = median of $kNN_{distances}$) on Wine UCI Dataset with three attributes red, blue and green nodes.}
        \label{fig:wineOur}
    \end{figure}

This building approach reduces the probability of non-connected graphs, capture the relations on the same scale reducing the dependency of normalization, and acquiring the hidden attribute-attribute patterns. We can observe the difference between the current technique described in equation (\ref{radius_network_construction_rule}), and our methodology on figure (\ref{fig:wineCurrent}) and figure (\ref{fig:wineOur}) respectively.


\section{Results}

In table \ref{table:resultAccuracy}, we present the results in normalized UCI datasets \cite{Dua:2019} with the current technique and our methodology. The dataset Wine presents attributes on different scales. For this reason, the current building methodology presents a reduced accuracy. However, our new methodology presents a better performance 95.56\% against 75.84\%. In the Iris dataset, the performance was similar in both technique due to this data set presents attribute with the same scale. In the last dataset zoo, our methodology reduces its performance drastically. This dataset has 17 attributes and 7 classes but just 100 instances. Theses characteristics could create noise in the final graph.

\begin{table}[h!]
\centering
\begin{tabular}{ |c|c|c|c| } 
 \hline
 \multicolumn{4}{|c|}{Results of 10 times using 10-folds cross validation} \\
\hline
Dataset & Prediction  & Building (k) & Accuracy \\

\hline
\multirow{2}{4em}{Iris} & \multirow{2}{4em}{HLNB\_BC} & 
kNN+$\epsilon$-$radius$ (1)  & 95.33 $\pm$ 11.85 \\ 
& &
Our methodology (2)  & 94.00 $\pm$ 10.41 \\ 
\hline
\multirow{2}{4em}{Wine} & \multirow{2}{4em}{HLNB\_BC} & 
kNN+$\epsilon$-$radius$ (1)  & 75.84 $\pm$ 19.15 \\ 
& &
Our methodology (1)  & 95.56 $\pm$ 10.18 \\ 
\hline
\multirow{2}{4em}{Zoo} & \multirow{2}{4em}{HLNB\_BC} & 
kNN+$\epsilon$-$radius$ (1)  & 96.36 $\pm$ 12.98 \\ 
& &
Our methodology (1)  & 60.09 $\pm$ 10.23 \\ 
\hline
% \hline
% \multirow{2}{4em}{Yeast} & \multirow{2}{4em}{HLNB\_BC} & 
% kNN+$\epsilon$-$radius$ (1)  & 89.45 $\pm$ 0.45 \\ 
% & &
% Our methodology (1)  & 89.45 $\pm$ 0.45 \\ 
% \hline
\end{tabular}
\caption{Table with accuracy of different building methodologies and UCI datasets without normalization. }
\label{table:resultAccuracy}
\end{table}

\section{Conclusion}
We introduce a new technique focused on capture the hidden patterns between attribute-attribute interactions.
The current results show us that capturing these patterns could improve the classification of high-level algorithms. However, this technique reduces its efficiency in small datasets with a high quantity of attributes and classes. 

\section{Future Works}

The proposed algorithm presents problems when the attribute-attribute relations do not provide additional information because they introduce noise to the final graph. Thus, it is needed a form to capture these attributes and remove their connections to avoid the noise.

\bibliographystyle{unsrt}  

% \bibliography{references}
%% Comment out this section when you \bibliography{references} is enabled.

\end{document}